# Two-Stage Convolutional Neural Network Architecture for Lung Nodule Detection

Haichao Cao, Hong Liu*, Enmin Song, Guangzhi Ma, Xiangyang Xu,

Renchao Jin, Tengying Liu, Chih-Cheng Hung

**Abstract**

Early detection of lung cancer is an effective way to improve the survival rate of patients. It is a critical step to have accurate detection of lung nodules in computed tomography (CT) images for the diagnosis of lung cancer. However, due to the heterogeneity of the lung nodules and the complexity of the surrounding environment, robust nodule detection has been a challenging task. In this study, we propose a two-stage convolutional neural network (TSCNN) architecture for lung nodule detection. The CNN architecture in the first stage is based on the improved UNet segmentation network to establish an initial detection of lung nodules. Simultaneously, in order to obtain a high recall rate without introducing excessive false positive nodules, we propose a novel sampling strategy, and use the offline hard mining idea for training and prediction according to the proposed cascaded prediction method. The CNN architecture in the second stage is based on the proposed dual pooling structure, which is built into three 3D CNN classification networks for false positive reduction. Since the network training requires a significant amount of training data, we adopt a data augmentation method based on random mask. Furthermore, we have improved the generalization ability of the false positive reduction model by means of ensemble learning. The proposed method has been experimentally verified on the LUNA dataset. Experimental results show that the proposed TSCNN architecture can obtain competitive detection performance.

# 1 Introduction

Lung cancer is one of the most dangerous diseases leading to cancer death,

accounting for two-thirds of all cancers [1, 2]. Its 5-year survival rate is 18% [3]. Clinical experience has shown that if lung cancer can be diagnosed at an early stage, the chance of survival will be greatly increased [4]. The use of diagnostic methods based on computed tomography (CT) images is an important strategy for early diagnosis of lung cancer and improvement of patient survival [5]. Accurate detection of lung nodules is an important step in the diagnosis of early stage lung cancer in medical imaging-based diagnostic methods. With an increasing number of CT images of lung nodules, in order to reduce the cumbersome manual labeling and the variability of detection results, it is of great clinical significance to develop robust automatic detection models [6].

In recent years, although many methods of lung nodule detection has been proposed [7-9], it is still difficult to obtain satisfactory detection result due to the heterogeneity of lung nodules on CT images (as shown in Fig. 1). For example, for calcific nodules (Fig. 1 ($tp_2$)), cavitary nodules (Fig. 1 ($tp_3$)) and ground-glass opacity nodules (Fig. 1 ($tp_4$)), they reflect the heterogeneity of lung nodules in terms of shape, texture and intensity. In addition, the development of a robust detection model is also a challenge due to the high degree of similarity between the lung nodules and their surrounding tissues. For example, for juxtapleural nodules (Fig. 1 ($tp_5$)), since the lung nodules are almost identical, in terms of intensity, to the lung wall, it is difficult to automatically locate its exact location. Similarly, for small lung nodules less than 6 mm in diameter such as Fig. 1 ($tp_6$) (corresponding to the green rectangle box) are difficult to distinguish because they have similar intensities to the surrounding noise which can be seen in Fig. 1 (fp6) (corresponding to the red rectangle box). Furthermore, to illustrate the similarity between lung nodules and non-lung nodules, we list six false positive lung nodules in the second row of Fig. 1 (Fig. 1 ($fp_{1-6}$)).

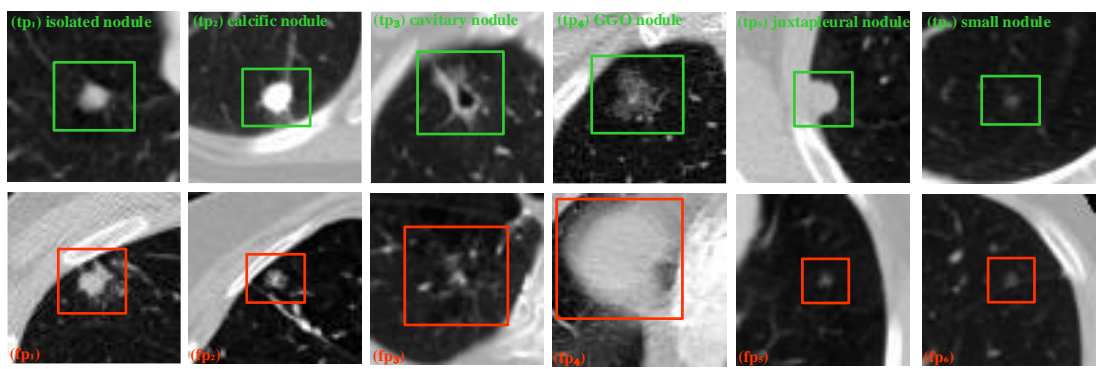

Fig. 1. Example images of lung nodules with different locations and shapes in CT image. Note that the GGO in sub-figure ($tp_4$) represents a ground-glass opacity nodule, and sub-figure ($tp_6$) is a small nodule with a diameter of 4.4 mm. Among them, the sub-figure ($fp_{1-6}$) represents the false positive nodule having similar appearance characteristics to the true lung nodule.

To solve the problems discussed above in the heterogeneous CT data, we propose a two-stage convolutional neural network (TSCNN) in this study. This network architecture is divided into two stages: the candidate nodule detection stage which is based on the improved UNet and the false positive reduction stage which is based on 3D CNN. The purpose of the first stage is to obtain a region of interest where a lung nodule may be present; the purpose of the second stage is to reduce false positives in candidate nodules obtained by the first stage. In general, TSCNN can detect various types of lung nodules to achieve a better detection rate. Our technical contributions in this work are the followings.

(1) A UNet segmentation network based on ResDense structure (Section 3.1.1) was designed and used to perform initial detection of lung nodules. In addition, we propose a new sampling strategy (Section 3.1.2) to select samples for training, and then train based on the offline hard mining idea (Section 3.1.3) to make the model suitable for those indistinguishable samples. Finally, using the proposed cascade prediction method (Section 3.1.5) for prediction can effectively reduce false positive nodules.

(2) Based on the design of the dual pooling approach, we have built three 3D CNN network architectures dedicated to reducing false positive lung nodules, which are based on SeResNet (Section 3.2.1), DenseNet (Section 3.2.2) and InceptionNet (Section 3.2.3) classification networks. It is worth noting that in order to obtain a better classification effect; we propose a data augmentation method based on random mask (Section 3.2.4). In addition, we have further improved the generalization ability of the proposed false positive reduction model by means of ensemble learning.

# 2 Related works

The detection of lung nodules usually involves two stages: one is the detection of candidate lung nodules and the other is the reduction of false positive lung nodules. In recent years, many solutions have been proposed. These methods can be generally divided into traditional detection methods, machine learning algorithms, and methods based on convolutional neural networks. Please note that the lung nodules will be referred to the nodules.

In the traditional detection method, in order to detect nodules from the very complex lung environment, morphological operations, threshold-based methods, clustering algorithms and energy optimization algorithms have been widely used [10-16]. For example, Gupta et al. developed a lung mask by combining masks, flooding fill algorithms, and morphological operations, and then performed nodule detection based on a multi-level thresholding algorithm combined with various feature extraction techniques [17]. Rezaie et al. first selected a region of interest that may have nodules based on the threshold method, and then used an edge detection algorithm to locate the nodule [18]. Typically, Lu et al. propose a hybrid algorithm for nodule detection that integrates traditional methods such as morphological operations, Hessian matrices, fuzzy sets, and regression trees [19].

In machine learning methods, researchers combine the classification models with advanced features for the detection of nodules [20-25]. For example, Froz et al. used artificial crawler and rose diagram techniques to extract texture features of nodules and then use radial basis kernel based support vector machines (SVM) for classification [26]. Similarly, Aghabalaei et al. designed a set of spectral, texture, and shape features to characterize nodules, and then used the SVM to classify candidate nodules [27]. Nithila et al. developed a Computer-Aided Detection (CAD) system for isolated lung nodule detection that focuses on heuristic search algorithms and uses particle clustering algorithms for network optimization [28]. In addition, Alam et al. proposed a patch-based multi-spectral method for nodules detection [29].

In the methods based on convolutional neural network (CNN), the researchers train the nodule detection model end-to-end in a supervised learning manner, while using the CNN to learn the relevant features of the nodule to replace traditional feature extraction methods. Within the CNN methods, the 2D CNN method has been widely used [30-38]. For example, Ciompi et al. used a combination of three views of the axial, sagittal, and coronal planes as input to the nodule detection model and used ensemble learning for prediction [39]. One year later, Setio et al. integrated the six diagonal views of the nodule into the input of the nodule detection model, each of which was processed by a ConvNets stream. The confidence of the candidate nodule is obtained by fusing the output of nine ConvNets which have the same structure [40]. George et al. used a YOLO-based method which is a method for object detection in natural images to detect lung nodules in CT images [41]. Meanwhile, some other researchers use 3D deep CNNs for the detection of nodules [42-46]. For example, Hamidian et al. use a fully convolutional network to generate negative samples that are difficult to be identified correctly, and then train a 3D CNN for nodule detection based on these negative samples [47, 48]. In order to simplify the detection process of nodules, Jenuwine et al. developed a CAD system that uses 3D CNN to detect nodules in CT images without using false positive reductions in candidate nodules, but the detection accuracy is far from expected [49]. Typically, a new 3D CNN with dense connections proposed by Khosrava et al., which is trained in an end-to-end manner, does not require any post-processing or user guidance to improve the detection results [50].

The TSCNN architecture proposed in this paper differs from the previous methods in the following aspects: 1) using the improved UNet segmentation model for lung nodule detection; 2) for the segmentation model training, we propose a new sampling strategy and an offline hard mining training approach; 3) we propose a cascade prediction method different from the traditional prediction method; 4) build three 3D CNN classification networks based on the dual pooling method; 5) design a data augmentation method based on random mask.

# 3. Methods

The lung nodule detection framework proposed in this paper is divided into two stages. The first stage: the detection of candidate nodules, which is based on the UNet architecture to achieve the detection of candidate nodules by segmenting suspicious nodules. The second stage: the reduction of false positive nodules, which is based on the 3DCNN architecture to eliminate false positive nodules through the integration of multiple models. The overall architecture of the proposed lung nodule detection method is shown in Fig. 2.

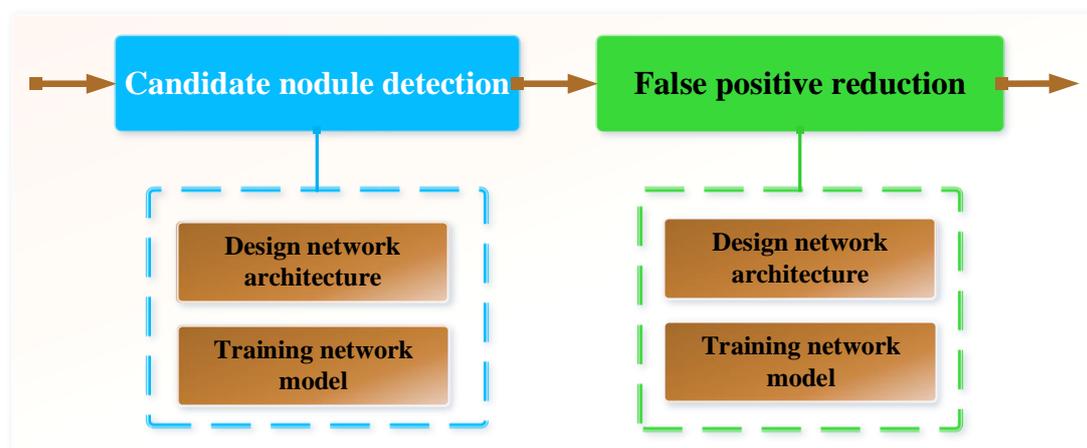

Fig. 2. The flowchart of the proposed two-stage CNN. Two stages are shown inside the blue and green boxes, respectively.

Section 3.1 describes the detection methods for candidate nodules, and Section 3.2 gives the false positive reduction methods for candidate nodules.

## 3.1. Candidate nodule detection

In this study, we use the UNet-based segmentation network to detect candidate nodules. In order to better detect candidate nodules, we improved the original UNet architecture and proposed a novel sampling strategy. In addition, to further reduce false positive nodules, we use the idea of the offline hard mining to train and use cascaded prediction methods for prediction.

### 3.1.1. Network architecture

Fig. 3 shows the proposed UNet segmentation architecture based on the residual-dense mechanism for initial lung nodule detection. The architecture consists of six residual dense blocks (ResDenseBlock1, ResDenseBlock2,..., ResDenseBlock6 in Fig. 3(a)), three 2D max pooling layers (three blue arrows in Fig. 3(a)), and three 2D deconvolution layers (three yellow arrows in Fig. 3(a)), and one sigmoid regression layer. Six residual dense blocks can be divided into two groups (each group contains three residual dense blocks), one for feature extraction including ResDenseBlock1, ResDenseBlock2, ResDenseBlock3, and the other for feature reconstruction including ResDenseBlock4, ResDenseBlock5, ResDenseBlock6. Each of the residual dense blocks described above is composed of the stack of six dense blocks, and one residual connection. Fig. 3(b) shows the basic unit-dense block constituting the residual dense block. The three 2D max pooling layers are separately used after three residual dense blocks in feature extraction. The three 2D deconvolution layers are used after three residual dense blocks in feature reconstruction. It should be noted here that the input of each residual dense block in feature reconstruction is the concatenation of the output of the upper layer and the output of its corresponding feature extraction layer.

In the specific experiment, the pooling kernel size, the stride and padding method of the three max pooling layers in Fig. 3(a) are 3×3, 2 and "same", respectively. The convolution kernel size, the stride and the padding method of the three deconvolution layers are the same as the max pooling layer. In addition, the number (i.e., n) of feature maps corresponding to different residual dense blocks in Fig. 3(b) is varied. The value of n corresponding to ResDenseBlock1 and ResDenseBlock6 is 8, the value of n corresponding to ResDenseBlock2 and ResDenseBlock5 is 16, and the value of n corresponding to ResDenseBlock3 and ResDenseBlock4 is 32.

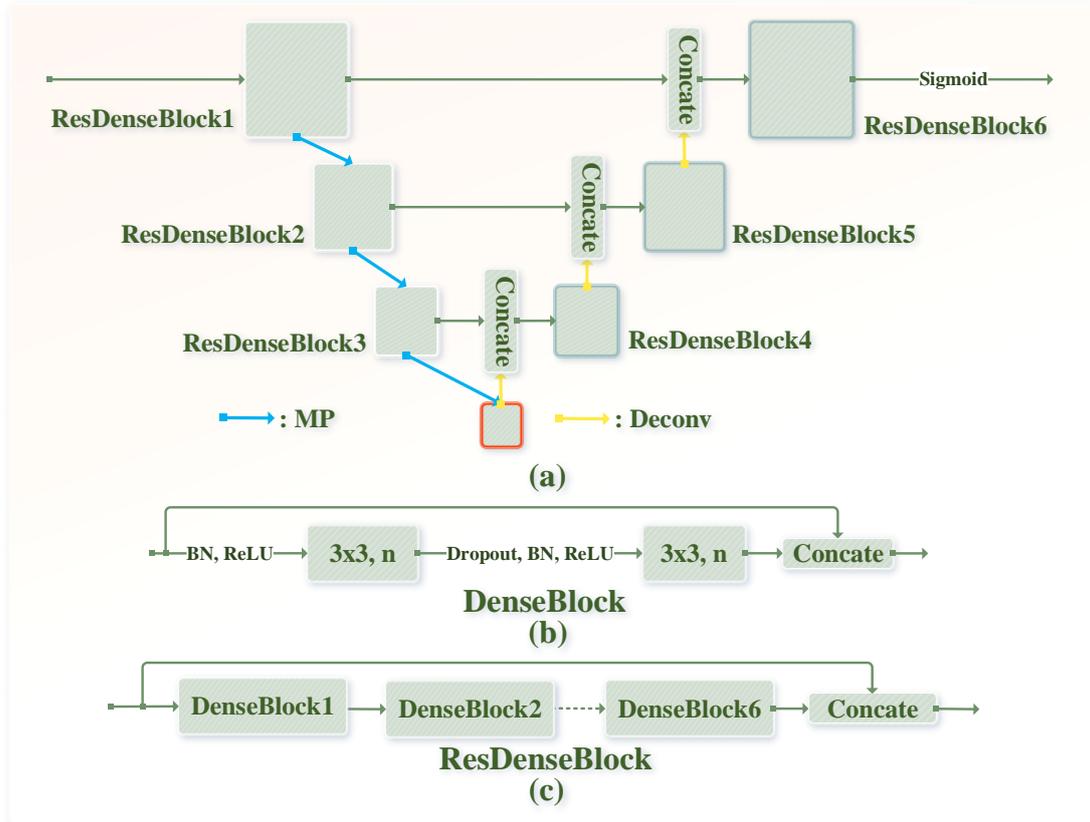

Fig. 3. (a) The proposed UNet segmentation architecture based on the residual-dense mechanism, (b) the diagram of the Dense block (DenseBlock), and (c) the diagram of the residual dense block (ResDenseBlock). The parameter "n" indicates the number of feature maps.

### 3.1.2. Sampling strategy

For the segmentation of lung nodules, the edge voxels of the nodules are crucial because they usually contain more texture information. If the edge voxels of the lung nodules are well recognized, a complete nodule mask can be obtained by simple morphological operation even if the internal segmentation of the nodules is incomplete. Based on this idea, when sampling the voxels of the lung nodule class, we consciously pay attention to the voxels on the edge of the lung nodules. Specifically, the corresponding sampling weight (the probability that a voxel is sampled) is set according to the minimum distance between each voxel and the lung nodule edge voxel. The smaller the distance is, the larger the weight will be. In addition, in order to have samples near the voxel which is close to the center of the nodule, we use the radius of the lung nodule as an adjustment factor to suppress the polarization of the sampling weight.

The calculation formula for the sampling weight of the voxel in the lung nodule class is shown in (1).

$$PW_k = \frac{\exp^{-\frac{\min_{t \subseteq E} d(k,t)}{r}}}{\sum \exp^{-\frac{\min_{t \subseteq E} d(k,t)}{r}}}, k \subseteq P \qquad (1)$$

where PW$_k$ represents the sampling weight of the k-th voxel in the lung nodule class; P the lung nodule class; E the set of voxels belonging to the edge of the lung nodule; d(k, t) the Euclidean distance between the k-th voxel in P and the t-th voxel in E, and r the radius of the lung nodule.

For the sampling of voxels in non-lung nodule class (such as blood vessels, lung walls, lung parenchyma, etc.), we divide them into two categories based on the distance from the currently sampled lung nodules. One class is the set of voxels that are closer to the currently sampled lung nodules (high correlation background class), and the corresponding sampling region is a local region (green region in Fig. 4) other than the lung nodules. The other class is the set of voxels that are farther from the currently sampled lung nodules (low correlation background class), and the corresponding sampling region is the region outside the high correlation background class sampling region (gray region of Fig. 4).

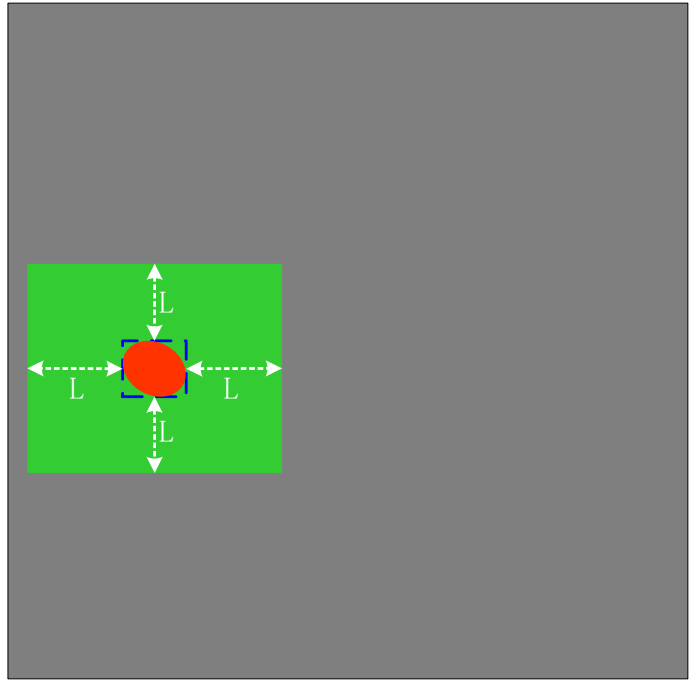

Fig. 4. The schematic diagram of the corresponding local region when sampling various voxels. The entire gray rectangle box represents one of the slices of the currently sampled nodule, and the red-labeled round (lung nodule) is a sampling region of the lung nodule; the blue dotted rectangular box representing the circumscribed rectangle of the lung nodule. In addition, the green rectangular region outside the lung nodule represents the sampling region of the high correlation background class, and the size of the green rectangular box is determined by the size of the lung nodule and the "L" value, where "L" indicates the length (the value in the experiment is 64) of the image patch used to train the UNet segmentation model. The gray region outside the green rectangle box indicates the sampling region of the low correlation background class.

In general, for the sampling of the voxels of non-lung nodule class, we consider their corresponding intensity information when calculating their sampling weights. This is because, compared with the lung parenchyma, air, and other low-intensity voxels, the lung wall, blood vessels, and the like with higher intensity are more similar to the lung nodules. Especially the juxtapleural nodules (the intensity of the adhered lung wall is almost the same as the intensity of the lung nodules). Therefore, for these indistinguishable voxels, a larger weight should be given, and the corresponding weight calculation formula is as shown in (2).

$$HBW_i = \frac{I_i \exp^{-\frac{\min_{j \subseteq E} d(i,j)}{r}}}{\sum I_i \exp^{-\frac{\min_{j \subseteq E} d(i,j)}{r}}}, i \subseteq HB \quad (2)$$

where $HBW_i$ represents the sampling weight of the i-th voxel in the high correlation background class; HB the high correlation background class; d(i, j) the Euclidean distance between the i-th voxel in HB and the j-th voxel in E, and $I_i$ the intensity value of the i-th voxel in HB.

For the voxels in low correlation background class that are far from the lung nodules, we consider the distance information between the voxels and the nodules in addition to the intensity information. The difference from the above is that the greater the distance from the center of the lung nodule (in order to speed up the sampling speed, where the distance from the edge of the lung nodule is not used) is, the greater the sampling weight will be. This is because the samples used to train the model should not all be voxels near the lung nodules, and some low correlation voxels should be sampled for training to improve the robustness of the model, thereby reducing false positive lung nodules. It should be noted that if the voxels sampled at this time belong to the lung nodule class, they should be eliminated. The calculation formula for the sampling weight of the low correlation background class is shown in (3).

$$LBW_p = \frac{I_p \times d(p,c)}{\sum I_p \times d(p,c)}, p \subseteq LB \quad (3)$$

where $LBW_p$ represents the sampling weight of the p-th voxel in the low correlation background class; LB the low correlation background class; c the central voxel of the currently sampled lung nodule; d(p,c) the Euclidean distance between the p-th voxel in LB and the voxel c, and $I_p$ the intensity value of the p-th voxel in LB.

Finally, we also need to determine the number of voxels that are sampled for each lung nodule. In order to adequately sample nodules of various sizes without introducing too many redundant samples, we only refer to the number of voxels on the edge of the nodule, not all voxels corresponding to the nodule. In the specific experiment, we set the total number of sampling points to 2.5 times the number of lung nodule edge voxels (N). In order to roughly balance various class of training samples, the number of sampling points of lung nodule class, high correlation background class and low correlation background class are N, N and 0.5N, respectively.

All of the above three sampling methods are to sample the slice containing the lung nodules. We know that the largest diameter of the lung nodules is 30mm (the lung nodules can span up to 30 slices), only a small part of the total number of slices (average of 300 slices), that is, many slices that do not contain lung nodules are not sampled which greatly reduces the generalization ability of the model. In the actual sampling process, if the slice containing no lung nodules is sampled according to the sampling rule of non-lung nodule class described above, or a simpler random sampling strategy is used, many redundant samples are inevitably introduced. Moreover, the scale of the training set will be expanded significantly, which will make the model difficult to converge, and will also greatly extend the training time of the model. To solve this problem, we do not directly sample the slice that does not contain the pulmonary nodules, but based on the idea of the offline hard mining, using the initial model M to predict the slice that does not contain the lung nodules, and then the samples of prediction error are treated as the sample sampled on the slice that does not contain a lung nodule, and finally added to the training set to continue training. The model M is the model obtained by training the sample sampled only from the slice containing the lung nodule.

### 3.1.3. Offline hard mining

As described in Section 3.1.2, in order to improve the robustness of the model and further reduce false positives, we indirectly sample the slices that do not contain lung nodules using the hard mining method. Specifically, we use the initial model M to predict the CT data in the training set. If any mask is segmented in the slice that does not contain the lung nodule, they will all be considered as difficult negative samples, and then their central voxels will be taken as our sampling points in the non-lung nodule slice. It should be noted that when predicting the slice containing the nodule (the segmentation mask does not intersect with the gold standard); the predicted result may also contain false positive lung nodules, which also need to be sampled. In addition, we still need to sample the lung nodule voxels to ensure the balance between the positive sample (the training sample containing some or all of the nodule mask in the label) and negative sample (the training sample that do not contain any nodule masks in the label) in retraining. However, for sampling, we are not randomly selecting voxels from the previously sampled lung nodules class that are equivalent to the number of negative samples. Instead, the number of samples of voxel of lung nodule class is dynamically adjusted based on the overlap rate between the current nodule and the gold standard. It is assumed that the number of edge points of the current nodule is C, and the overlap rate [51] with the gold standard is O, then the number of voxels sampled for this nodule is T=C*(1-O). The meaning of this formula is that the better the segmentation of the current nodule is, the fewer the number of corresponding sampling points will be. On the contrary, it indicates that the segmentation result of the model on the current nodule is not good, so it is necessary to pay attention to it, and the number of corresponding sampling points should also be increased.

### 3.1.4. Training procedure

In the actual training process, we crop the 64*64*3 size data patch centered on the voxels sampled from the slice containing the nodules for training the initial segmentation model M. When the training is completed, the model M is used to select difficult positive and negative samples, and then fine-tuned based on the initial model M to improve the generalization ability of the model. It should be noted that in the experiment, we use the Adam optimizer [52] to update the model parameters. In order to prevent over-fitting, we adopted a training strategy for early stop [53]. Specifically, if the performance of the model is not improved, the training continues for 5 epochs, and the total training algebra is 15 epochs. In addition, our initial learning rate is 0.0001, the learning rate when fine-tuning is 0.00001, and the batch size is 64.

The loss function is calculated as shown in equation (4), which differs from the dice coefficient in [54]. To allow the negative samples (V(Gt) is zero) and positive samples to be trained together, we add a non-zero parameter to the numerator and denominator of the original dice formula. Thus, even if V(Gt) and V(Seg) are both zero, it does not cause the denominator in equation (4) to be zero.

$$Dice = 1 - \frac{2 \times V(Gt \cap Seg) + \eta}{V(Gt) + V(Seg) + \eta} \qquad (4)$$

where V is the size of the volume, Gt the gold standard (mask label), and Seg the result of the model segmentation. In the experiment, the value of $\eta$ is set to 1.0.

### 3.1.5. Cascaded prediction

When the model training is completed, we predict the entire CT image by sliding the data patch of size 128*128*3 (stride size 64). For overlapping regions, in order to ensure a high recall rate, we use the union of the predicted results as the final segmentation result, but this will introduce more false positives nodules. To alleviate this problem, we take the result of this sliding prediction as the center, extract 64*64*3 data blocks, and send them into the model for secondary prediction. The purpose of this is to make an accurate segmentation in the local region where the suspected nodule is located. Because even if the suspected nodule is not a true lung nodule, narrowing the prediction area to the size of the data patch used during training will be more conducive to model prediction.

Table 1. Results of the cascading predictions on the testing set. Among them, The FPS indicates false positive rate and SEN shows sensitivity.

|                  | FPS    | Sen   |
|------------------|--------|-------|
| First Prediction | 139.40 | 0.988 |
| Second Prediction | 37.83 | 0.954 |

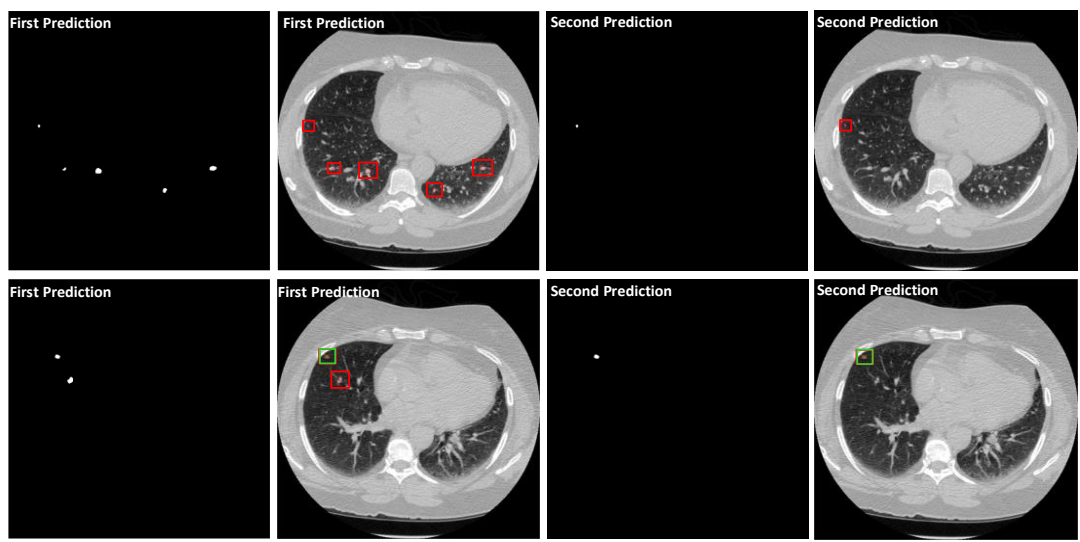

Fig. 5. Segmentation results for the first stage segmentation model in layers that do not contain lung nodules (first row) and layers that contain lung nodules (second row). The first and second prediction results are drawn in the second column and the fourth column, where red indicates the rectangular box corresponding to the model segmentation result, and green specifies the rectangular box corresponding to the gold standard.

To verify the effect of the cascade prediction, we only evaluate the results of the first stage using the segmentation model for nodule detection. The evaluation results are shown in Table 1. As can be seen from Table 1, the false positive rate has dropped nearly fourfold when the recall rate has only decreased by less than four percentage points. In addition, in order to more intuitively observe the necessity of the second prediction, we randomly selected the segmentation results of the two-layer slice in a certain CT data in the test set. As shown in Fig. 5, the first row is the slice that does not contain the lung nodules, and the second row is the slice that contains the lung nodules. As can be seen from the results shown in Figure 5, in the slice containing no nodules, the false positive nodules were greatly reduced after the second prediction; when predicted in the slice containing nodules, the false positives were also reduced, and the true positive nodules were not missed.

## 3.2. False positive reduction

Our proposed 3DCNN-based false positive reduction module contains three network models, which are based on the 3D dual pooling network architecture of SeResNet [55], DenseNet [56] and InceptionNet [57]. For the center of a given candidate lung nodule, we extract a 3D data patch of size 40*40*40 containing the lung nodule as input to the three network models, and by combining the output probabilities of the three models to obtain the final prediction results. Fig. 6 shows an overall architectural diagram of the proposed method of reducing false positive lung nodules.

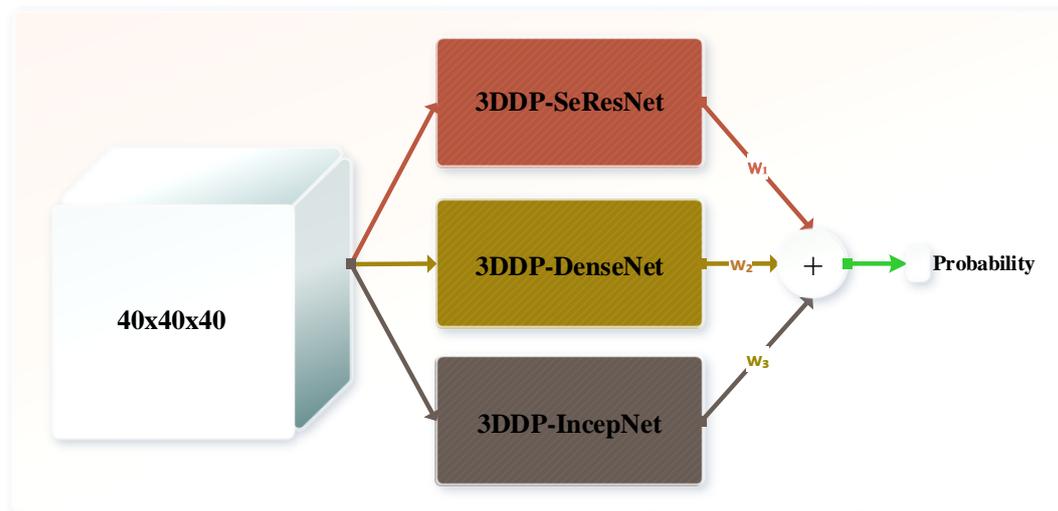

Fig. 6. The proposed 3D CNN-based ensemble learning framework. The 3DDP-DenseNet, 3DDP-SeResNet and 3DDP-IncepNet represents 3D Dual Pooling DenseNet, SeResNet and InceptionNet, respectively. The final prediction result is a weighted average of the output probabilities of the three modules, where w1, w2, and w3 are all one-third.

### 3.2.1. 3D Dual Pooling SeResNet

Fig. 7 shows the architecture of the proposed 3D dual pooling SeResNet

(3DDP-SeResNet) and Table 2 lists the corresponding network parameters.

The network consists of 38 convolutional layers, three dual pooling layers (the combination of the central pooling layer [58] and the central clipping layer, referred to as DP) and one global average pooling layer. The 38 convolutional layers of the network are divided into two categories, one is a convolution block composed of two stacked convolution layers (as shown in Fig. 7(b)), and the other is a SeResBlock group composed of three stacked SeResBlock (as shown in Fig. 7(d)).

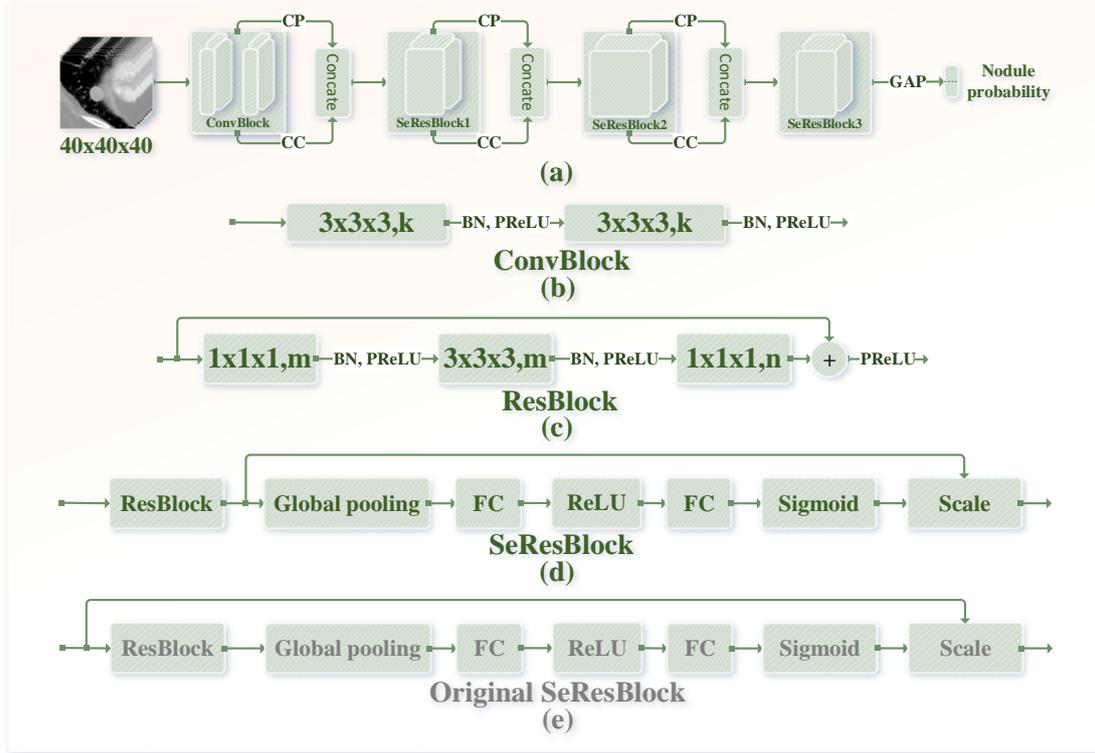

Fig. 7. (a) The proposed 3D Dual Pooling SeResNet (3DDP-SeResNet), where the symbols "CP" "CC" and "GAP" represent the Central Pooling, the Central Cropping and the Global Average Pooling, respectively. (b) The structural diagram of 3D convolution block (ConvBlock) in which the symbols "k", "BN" and "PReLU" represent the number of channels, the batch normalization operation, and the nonlinear activation function, respectively. (c) The structural diagram of 3D residual block (ResBlock) in which the symbols m and n represent the number of channels. (d) The structural diagram of the proposed SeResBlock. (e) The structural diagram of the original SeResBlock.

Table 2. It shows network parameters of the 3DDP-SeResNet. Building blocks are shown in brackets with the numbers of blocks stacked. Downsampling is performed using Dual Pooling before the first layer of SeResBlock1, SeResBlock2 and ResBlock3. The stride size of the convolution operation is one. The symbol "*" indicates that there is no such operation.

| Layer name | 3DDP-SeResNet |
| --- | --- |
| Input size | $40 \times 40 \times 40$ |
| ConvBlock | $[3 \times 3 \times 3,\ 32] \times 2$ |
| SeResBlock1 | $\begin{bmatrix} conv,\ 1\times1\times1,\ 32 \\ conv,\ 3\times3\times3,\ 32 \\ conv,\ 1\times1\times1,\ 64 \\ fc,\ [8,\ 64] \end{bmatrix} \times 3$ |
| SeResBlock2 | $\begin{bmatrix} conv,\ 1\times1\times1,\ 48 \\ conv,\ 3\times3\times3,\ 48 \\ conv,\ 1\times1\times1,\ 96 \\ fc,\ [12,\ 96] \end{bmatrix} \times 6$ |
| SeResBlock3 | $\begin{bmatrix} conv,\ 1\times1\times1,\ 64 \\ conv,\ 3\times3\times3,\ 64 \\ conv,\ 1\times1\times1,\ 128 \\ fc,\ [16,\ 128] \end{bmatrix} \times 3$ |
| Output size | 5×5×5 |
| Nodule probability | Global Average Pooling, 2-d fc, softmax |

To avoid the disappearance of the gradient and speed up the convergence, the batch normalization operation [59] is used. After each convolution, we use the nonlinear parameter rectification linear unit (PReLU) as the activation function [60]. It should be noted that the SeResBlock we designed here is slightly different from the original paper [55]. That means we use the features extracted by the residual block as input to the Squeeze-and-Excitation mechanism instead of integrating them together (as shown in Fig. 7(d)). Experiments have shown that this structure is superior to the original SeResBlock structure (as shown in Fig. 7(e)). Meanwhile, we replaced the downsampling operation in [55] with the proposed DP operation.

### 3.2.2. 3D Dual Pooling DenseNet

Fig. 8 shows the architecture of the proposed 3D dual pooling dense network (3DDP-DenseNet), and Table 3 lists its corresponding network parameters. The network consists of 13 convolutional layers, three DP layers and one global average pooling layer. Among them, the 13 convolutional layers of the network are divided into two categories; one is a convolution block composed of two stacked convolution layers (as shown in Fig. 7(b)), and the other is a dense block group composed of three stacked dense blocks (as shown in Fig. 8(b)). In addition, similar to 3DDP-SeResNet, 3DDP-DenseNe replaces the max pooling operation of the original paper [56] using DP operations. The rest of the operations, such as batch normalization, the type of activation function, and the fully connected layer, are similar to those used in 3DDP-SeResNet.

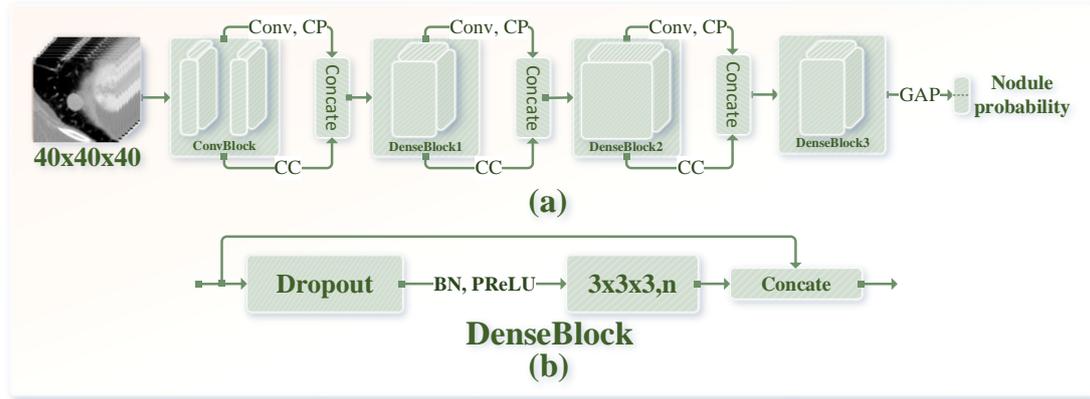

Fig. 8. (a) The proposed 3D Dual Pooling DenseNet (3DDP-DenseNet). (b) The diagram of the 3D dense block (DenseBlock). The parameter n indicates the number of channels. The "CP", "GAP" and "Concate" represent the Central Pooling, Global Average Pooling and Concatenate operation, respectively. The ConvBlock is the same as described in Fig. 7(b).

Table 3. Network parameters of the 3DDP-DenseNet are listed in the following. The growth rate and dropout rate for the networks are 32 and 0.2, respectively.

| Network name | 3DDP-DenseNet |
| --- | --- |
| Input size | $40 \times 40 \times 40$ |
| ConvBlock | $[3 \times 3 \times 3,\ 32] \times 2$ |
| Transition Layer | $1 \times 1 \times 1$ Conv, Central Pooling, Central Cropping, Concatenate |
| DenseBlock1 | $\begin{bmatrix} \text{Dropout} \\ 3 \times 3 \times 3\ \text{Conv} \end{bmatrix} \times 2$ |
| Transition Layer | $1 \times 1 \times 1$ Conv, Central Pooling, Central Cropping, Concatenate |
| DenseBlock2 | $\begin{bmatrix} \text{Dropout} \\ 3 \times 3 \times 3\ \text{Conv} \end{bmatrix} \times 4$ |
| Transition Layer | $1 \times 1 \times 1$ Conv, Central Pooling, Central Cropping, Concatenate |
| DenseBlock3 | $\begin{bmatrix} \text{Dropout} \\ 3 \times 3 \times 3\ \text{Conv} \end{bmatrix} \times 2$ |
| Output size | $5 \times 5 \times 5$ |
| Nodule probability | Global Average Pooling, 2-d fc, softmax |

### 3.2.3. 3D Dual Pooling IncepNet

Fig. 9 shows the architecture of the proposed 3D dual pooling inception network (3DDP-IncepNet), and Table 4 lists its corresponding network parameters. As shown in Figure 9, the network contains 41 convolutional layers, three DP layers, and one global average pooling layer. Among them, the 41 convolutional layers of 3DDP-IncepNet are divided into two categories; one is a convolution block composed of two stacked convolutional layers shown in Fig. 7(b), and the other is an inception block group composed of three stacked inception blocks shown in Fig. 9(b). In addition, the inception block we designed is different from the Inception structure in [57]. The main difference is that we incorporate central pooling and center cropping into the inception structure to better extract multi-scale features of lung nodules.

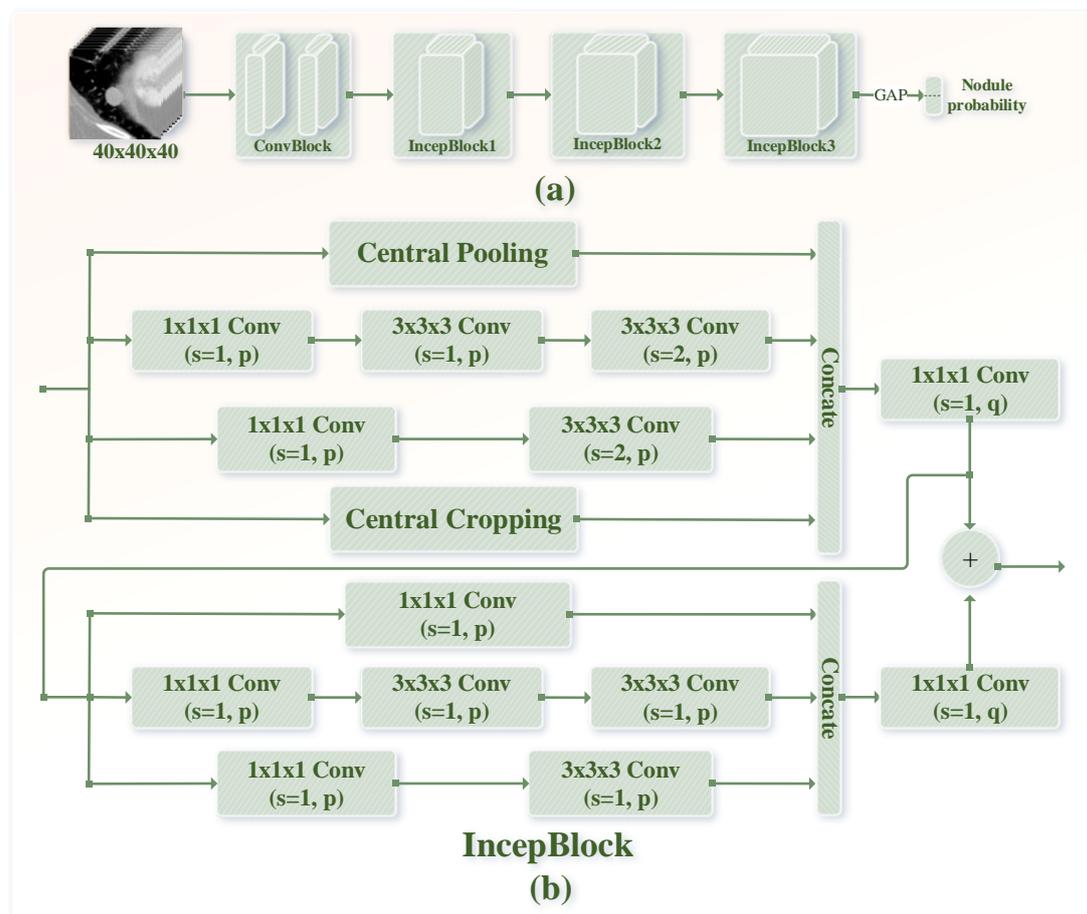

Fig. 9. (a) The proposed 3D Dual Pooling InceptionNet (3DMB-IncepNet). (b) The diagram of the 3D Inception block (IncepBlock). The parameters p and q indicate the number of channels. The "GAP" and "Concate" represent the Global Average Pooling operation and Concatenate operation, respectively. The ConvBlock is the same as described in Fig. 7(b).

Table 4. Network parameters of the 3DDP-IncepNet.

| Layer name | 3DDP-IncepNet |
|---|---|
| Input size | $40 \times 40 \times 40$ |
| ConvBlock | $[3 \times 3 \times 3,\ 32] \times 2$ |
| IncepBlock1 | p=16, q=64 |
| IncepBlock2 | p=24, q=96 |
| IncepBlock3 | p=32, q=128 |
| Output size | $5 \times 5 \times 5$ |
| Nodule probability | Global Average Pooling, 2-d fc, softmax |

### 3.2.4. Random mask

To ensure a high recall rate, when using the segmentation model to search the candidate nodules in the first stage, it is inevitable that some false positive nodules will be introduced. In order to ensure the reasonableness of the evaluation results, when training the false positive reduction model, we only use the candidate nodule obtained in the first stage as the training sample, and do not use the annotation file provided by the second track of luna16 (because it contains some samples detected in the first stage). Even if the sample obtained in the first stage is used to train the false positive reduction model, there is an imbalance problem on the number of positive and negative samples. To solve the problem of the imbalance on the number of positive samples (true lung nodules) and negative samples (blood vessels, lung walls, lung parenchyma, etc.), we expand the data of positive samples in the training set to make the ratio of positive and negative samples tends to be 1. Among them, the expansion method is to rotate the lung nodules by 90°, 180° and 270° on the axial plane, and to translate one voxel in the X, Y and Z directions. Later, in order to further improve the generalization ability of the model, we extend the positive and negative samples in the training set by the random mask data augmentation method. Specifically, the expansion for positive samples is to randomly select T (5% of the number of negative samples) samples for the expanded positive and negative samples. Then, the selected T positive samples and T negative samples are randomly paired; finally, centering on the nodule, the data patch of size D*D*D (D is the diameter of the current nodule) is cropped to replace the data patch corresponding to the negative sample paired with it, thereby converting the negative sample into a new positive sample. The expansion of the negative samples is relatively simple, that is, only T samples need to be randomly selected from the positive samples, and then the corresponding data patch of size D*D*D is set to 0 with the nodule as the center (i.e., set the intensity of each voxel in the data patch to 0), so that the T positive samples are converted into T negative samples.

### 3.2.5. Training procedure

The training procedure for the three 3D CNN network architectures for false positive reduction is identical. The SGD optimizer is used to update the model parameters, and the corresponding initial learning rate is 0.001. After that, each generation of learning rate decays to 90% of the previous generation learning rate. In addition, their corresponding total training algebra, batch size, and momentum are 20, 64, and 0.9, respectively. Finally, we maximize the probability of the correct class by minimizing the cross entropy loss of each training sample. For an input sample given a positive or negative label, assuming that y is its true label, then the loss function is defined as shown in equation (5):

$$L = -\frac{1}{N}\sum_{n=1}^{N}[y_n log(y_n') + (1-y_n)log(1-y_n')] \quad (5)$$

where y' is the predicted probability of the model and N is the number of samples.

# 4. Data and experiment

Below we will further explain the data used and related experiments. The chapter is divided into three subsections, Section 4.1 describes the data used, Section 4.2 describes the evaluation criteria, and Section 4.3 describes the experimental environment.

## 4.1. Data

We used the data set containing 1186 lung nodules provided by the first phase of LUNA16 (the annotation information includes the diameter and location information of the nodules) [61]. Furthermore, it should be noted that this dataset is a subset of the public LIDC-IDRI data set containing 2610 lung nodules. They screened 888 CT data from the LIDC-IDRI dataset, each of which was labeled by up to four experienced radiologists [62]. Moreover, each radiologist classifies the identified lesions into three categories, non-nodular (other tissues or background), nodules larger than 3 mm in diameter, and nodules less than 3 mm in diameter. Finally, nodules larger than 3 mm in diameter, marked by three or four radiologists, are used as the gold standard, and nodules that are less than 3 mm in diameter and marked by only one or two radiologists will be ignored.

## 4.2. Evaluation criteria

Our evaluation criteria are the same as those used in the LUNA16 competition. The competition performance metric (CPM) was defined as the average sensitivity of seven predefined false positive rates (these seven values are 0.125, 0.25, 0.5, 1, 2, 4, and 8), and calculated as follows.

$$CPM = \frac{1}{N}\sum_{i=\{0.125,0.25,0.5,1,2,4,8\}} Recall_{fpr=i} \quad (6)$$

wherein, the value of "N" is seven, "fpr" is the average number of false positives per scan, and "Recall$_{fpr=i}$" is the recall rate corresponding to fpr=i.

## 4.3. Experimental environment

The hardware environment for all our experiments is the server with an Intel(R) Xeon(R) processor and 125GB of RAM. Moreover, the server has 10 GPUs, the model is GTX-1080Ti GPU, and the video memory size is 11GB. In addition, the software environment on which all experiments are based is operating system ubuntu 14.04, integrated development tool PyCharm 2017.2.1, coding language Python 3.6.4, deep learning framework Keras. According to our experiments, the time required for the 3DDP-SeResNet, 3DDP-DenseNet and 3DDP-IncepNet models to converge was approximately 18, 16 and 13 hours, respectively.

# 5. Results and discussion

This section is divided into three subsections, of which Section 5.1 describes the overall performance of the proposed method, Section 5.2 describes the ablation study of the proposed method, and Section 5.3 describes our comparative experiment.

## 5.1. Overall performance

To more intuitively observe the overall performance of the proposed false positive reduction method, we plot the FROC curves for 3DDP-DenseNet, 3DDP-SeResNet and 3DMB-InceptionNet, and their integrated corresponding FROC curves in Fig. 10. In these three networks, when the false positive rate of each scan is 1, the corresponding sensitivity can reach more than 90%; their integration results in a false positive rate of 0.25, the corresponding sensitivity reached 90%, which indicates that the network architecture based on 3D CNN can adapt to the classification of true and false lung nodules to some extent.

Table 5 lists the quantitative metrics for the three independent network architectures, namely the sensitivity at the different false alarm rates specified in the Challenge. As can be seen from Table 5, when the false positive rate of each scan is 0.125, the recall rates of the three networks proposed are less than 80%, and their integration corresponds to the sensitivity of 84.8%, which is compared with 3DDP-IncepNet-RDU increased by 15.8 percentage points. At this extremely low false positive rate, our multi-model integration strategy has shown satisfactory results.

Even for the final CPM score, the integrated result is 2.5 percentage points higher than the best-performing 3DDP-SeResNet-RDU. These experiments show that different network architectures can complement each other's defects, and their integration can improve the performance.

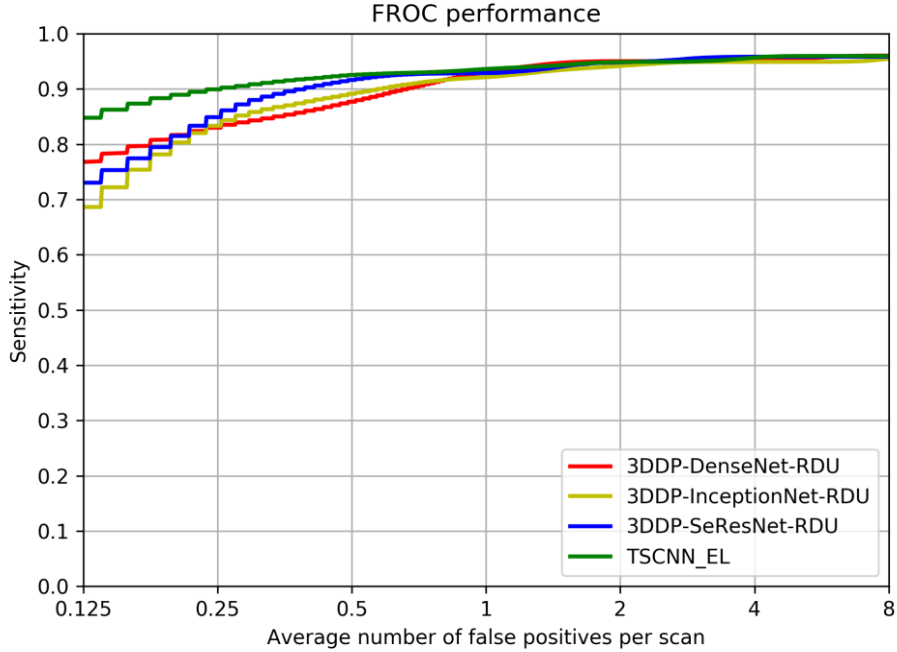

Fig. 10. FROC curves for the three networks and their integration.

Table 5. The sensitivities of models under different false positive rates. Note that, RDU indicates ResDense-based UNet architecture and uses the proposed sampling strategy to train the segmentation model.

| Method | 0.125 | 0.25 | 0.5 | 1 | 2 | 4 | 8 | CPM |
|---|---|---|---|---|---|---|---|---|
| 3DDP-DenseNet-RDU | 0.768 | 0.830 | 0.877 | 0.931 | 0.950 | 0.951 | 0.960 | 0.895 |
| 3DDP-SeResNet-RDU | 0.730 | 0.849 | 0.916 | 0.929 | 0.948 | 0.958 | 0.958 | 0.900 |
| 3DDP-IncepNet-RDU | 0.686 | 0.833 | 0.891 | 0.921 | 0.942 | 0.949 | 0.955 | 0.883 |
| TSCNN | 0.848 | 0.900 | 0.925 | 0.936 | 0.949 | 0.957 | 0.960 | 0.925 |

## 5.2. Ablation study

To validate the effectiveness of the various components in the TSCNN architecture, we designed a 2D DenseNet-based ablation experiment. The results are shown in Table 6. It should be noted that in order to shorten the length, we only randomly select one fold for the ablation experiment, and the other tables show the results of the 10-fold cross-validation (except Tables 8, 9 and 10). In addition, the first four rows in Table 6 (except the header) verify the validity of the components in the proposed segmentation network, and the last seven rows verify the validity of the components in the proposed classification network. This will be explained below.

(1) The effectiveness of the sample selection

Comparing the first two rows of Table 6, it can be found that in the case of using 2D DenseNet for false positive reduction, the CPM score corresponding to the weighted sampling strategy [58] is 0.546, and the CPM score corresponding to the proposed sampling strategy is 0.563. The proposed sampling strategy increased performance by 1.7 percentage points, which validated the effectiveness of our proposed sampling strategy.

(2) The effectiveness of the basic structure

Comparing the second and third rows of Table 6, we can find that the UNet network model based on ResDense structure is 1.2% higher than the original UNet network architecture [63]. Especially, when the false positive rate is 0.125, the corresponding recall rate is particularly effective, which is 4.6 percentage points higher than the original UNet structure. This illustrates that the segmentation model based on ResDense structure is effective.

(3) The effectiveness of the hard mining

Comparing the third and fourth rows of Table 6, we can find that the segmentation model is trained based on the idea of offline hard Mining. Although the corresponding recall rate was lower than the conventional training method when the false positive rate was 0.125 and 0.25, the overall performance was improved by 2 percentage points. In general, the segmentation model based on HM idea training is acceptable.

(4) Effect of the dimensional space

As can be seen from the 4th and 5th rows of Table 6, in the case of ensuring that the same segmentation model is used, the classification model based on 2D DenseNet has a CPM score of 0.595, and the classification model based on 3D DenseNet has a CPM score of 0.835. By comparing these two results, it can be found that the classification model based on 3D DenseNet improves the performance by 24%. Although the model parameters are increased (3D DenseNet's trainable model parameters are three times that of 2D DenseNet), the performance is greatly improved, which verifies the effectiveness by using more spatial information.

(5) The effectiveness of the downsampling method

Comparing 5th and 6th rows of Table 6, it can be seen that the downsampling method based on the central pooling is slightly better than the traditional max pooling method. After we combine the central pooling with the center cropping to form the dual pooling for downsampling, the performance is greatly improved (the result corresponds to line 7 of Table 6). Compared with the traditional max pooling, the CPM score increased by 3.6 percentage points, which verified the effectiveness of the dual pooling of our design in our experiments.

(6) The effectiveness of the random mask

Comparing 7th and 8th rows of Table 6, it can be seen that after using the proposed data augmentation method based on random mask, the recall rate corresponding to the seven false positive rates was improved compared to the results in row 7 of Table 6. Although the final CPM score improvement effect is not significant (false positive rate 0.125 corresponding recall rate improvement effect is more obvious), but also 1.1 percentage points increase, which to some extent also verified the effectiveness of the proposed random mask based data augmentation method.

In addition, the effectiveness of ensemble learning can also be verified by comparing the last four lines of Table 6.

Table 6. Ablation experiment. Note that, "2DDNet" indicates the 2D-DenseNet; "MP" indicates max pooling; "WSS" represents weighted sampling strategy[58]; "PSS" represents the proposed sampling strategy; "HM" represents the hard mining operation, see section 3.1.3; "CP" indicates central pooling; "DP" represents the combination of central pooling and central cropping; "RM" represents the data enhancement strategy based on random mask. In addition, for the convenience of writing in Table 6, we have abbreviated "3D-DenseNet", "3D-SeResNet" and "3D-IncepNet" as "3DDNet", "3DSNet" and "3DINet", respectively.

| Method | 0.125 | 0.25 | 0.5 | 1 | 2 | 4 | 8 | CPM |
|---|---|---|---|---|---|---|---|---|
| 2DDNet_MP-UNet_WSS | 0.308 | 0.398 | 0.491 | 0.549 | 0.642 | 0.688 | 0.748 | 0.546 |
| 2DDNet_MP-UNet_PSS | 0.347 | 0.405 | 0.463 | 0.548 | 0.651 | 0.761 | 0.768 | 0.563 |
| 2DDNet_MP-RDU | 0.393 | 0.426 | 0.465 | 0.568 | 0.656 | 0.725 | 0.793 | 0.575 |
| 2DDNet_MP-RDU_HM | 0.262 | 0.379 | 0.480 | 0.625 | 0.750 | 0.801 | 0.866 | 0.595 |
| 3DDNet_MP-RDU_HM | 0.583 | 0.769 | 0.833 | 0.883 | 0.914 | 0.931 | 0.932 | 0.835 |
| 3DDNet_CP-RDU_HM | 0.654 | 0.748 | 0.835 | 0.882 | 0.912 | 0.916 | 0.931 | 0.840 |
| 3DDNet_DP-RDU_HM | 0.745 | 0.822 | 0.872 | 0.902 | 0.916 | 0.916 | 0.926 | 0.871 |
| 3DDNet_DP_RM-RDU_HM | 0.774 | 0.837 | 0.877 | 0.910 | 0.916 | 0.929 | 0.931 | 0.882 |
| 3DSNet_DP_RM-RDU_HM | 0.703 | 0.855 | 0.910 | 0.912 | 0.925 | 0.928 | 0.928 | 0.880 |
| 3DINet_DP_RM-RDU_HM | 0.730 | 0.846 | 0.877 | 0.898 | 0.923 | 0.934 | 0.934 | 0.877 |
| TSCNN | 0.868 | 0.900 | 0.913 | 0.915 | 0.916 | 0.931 | 0.932 | 0.911 |

## 5.3. Experimental comparison

To verify the effectiveness of the proposed methods, we compared them with the nodule detection approaches published in recent years. Table 7 shows the quantitative results in the comparison of traditional detection methods, machine learning based detection methods, and CNN-based detection methods. It can be seen from Table 7 that although the detection methods based on the nodule enhancement filter and the support vector machine proposed by Teramoto et al. [24] are not effective, the machine learning-based detection methods and CNN-based detection methods are generally superior to traditional detection methods. In addition, for traditional machine learning-based detection methods and CNN-based detection methods, we believe that CNN-based deep learning methods are more scalable and robust. Generally speaking, some machine learning-based detection methods have achieved good results. For example, Liu et al. [25] and Javaid et al. [21] proposed methods based on traditional machine learning algorithm for lung nodule detection, but they need to manually design related features for later classification, which greatly limits the scalability of the method. However, the CNN-based approach is more natural, which automatically learns advanced semantic features through training. The methods we proposed are based on the CNN. By comparing the experimental results in Table 7, it is competitive with existing lung nodule detection methods.

Table 7. A comparison on the quantitative results of various false positive reduction methods. Among them, "Ours" indicates the result of the 10-fold cross-validation of our proposed method. Please note that, "TA" indicates the traditional algorithm, "ML" indicates the machine learning, and "CNN" indicates the convolution neural networks. In addition, "V$^-_{fps=n}$" means that when there are n false positives per scan, the corresponding recall rate is smaller than the value V. For example, the corresponding results of El-Regaily (V=0.705$^-$, n=4) indicate that when there are 4 false positives per scan, the corresponding recall rate is less than or equal to 0.705 (The original paper [15] only shows that when there are 4.1 false positives per scan, the corresponding recall rate is 0.705).

| Type | Method | 0.125 | 0.25 | 0.5 | 1 | 2 | 4 | 8 | CPM |
|---|---|---|---|---|---|---|---|---|---|
| TA | El-Regaily [15] | * | * | * | * | * | 0.705$^-$ | * | * |
|  | Anoop [16] | * | * | * | * | * | 0.850$^-$ | * | * |
|  | Lu [19] | * | * | * | * | 0.852$^-$ | * | * | * |
|  | Wang [12] | * | * | * | * | * | 0.880 | * | * |
| ML | Liu [23] | * | * | * | 0.893$^-$ | * | * | * | * |
|  | Teramoto [24] | * | * | * | * | * | * | 0.830$^-$ | * |
|  | Liu [25] | * | * | * | * | * | 0.948$^-$ | * | * |
|  | Javaid [21] | * | * | * | * | 0.917$^-$ | * | * | * |
| CNN | Hamidian [47] | 0.583 | 0.687 | 0.742 | 0.828 | 0.886 | 0.918 | 0.933 | 0.797 |
|  | Xie [37] | 0.734 | 0.744 | 0.763 | 0.796 | 0.824 | 0.832 | 0.834 | 0.790 |
|  | Dou [38] | 0.659 | 0.745 | 0.819 | 0.865 | 0.906 | 0.933 | 0.946 | 0.839 |
|  | Zhu [45] | * | * | * | * | * | * | * | 0.842 |
|  | Ding [46] | * | * | * | * | * | * | * | 0.891 |
|  | Khosravan [50] | * | * | * | * | * | * | * | 0.897 |
| **CNN** | **Ours** | **0.848** | **0.899** | **0.925** | **0.936** | **0.949** | **0.957** | **0.960** | **0.925** |

# 6. Conclusion

In this study, we designed a two-stage convolutional neural network architecture to better detect lung nodules. In general, we first use the proposed UNet segmentation model based on ResDense structure to search suspicious nodules (the centroid of the segmented mask is the location of the suspected lung nodules), and then use the proposed 3D CNN-based ensemble learning architecture to eliminate false positive nodules. In addition, we verified each component and overall performance of the proposed lung nodule detection method by ablation study and experimental comparison. According to the experimental results shown in Tables 6 and 7, we can see that our results are competitive compared to other existing technical methods.

# Acknowledgements

The National Key R&D Program of China (Grant No. 2017YFC0112804) and the National Natural Science Foundation of China (Grant No. 81671768) supported this work. The authors acknowledge the National Cancer Institute and the Foundation for the National Institutes of Health and their critical role in the creation of the free publicly available LIDC-IDRI Database used in this study.